# Point-Based POMDP Algorithms: Improved Analysis and Implementation

**Trey Smith and Reid Simmons**
Robotics Institute, Carnegie Mellon University
Pittsburgh, PA 15213

## Abstract

Existing complexity bounds for point-based POMDP value iteration algorithms focus either on the curse of dimensionality or the curse of history. We derive a new bound that relies on both and uses the concept of discounted reachability; our conclusions may help guide future algorithm design. We also discuss recent improvements to our (point-based) heuristic search value iteration algorithm. Our new implementation calculates tighter initial bounds, avoids solving linear programs, and makes more effective use of sparsity. Empirical results show speedups of more than two orders of magnitude.

## 1 INTRODUCTION

Partially observable Markov decision processes (POMDPs) constitute a powerful probabilistic model for planning problems that include hidden state and uncertainty in action effects. Recently, several POMDP solution algorithms have been developed that use approximate value iteration with point-based updates. These algorithms have proven to scale very effectively, relying on the fact that performing many fast approximate updates often results in a more useful value function than performing a few exact updates.

Point-based updates are applied over a set $\mathcal{B}$ of beliefs drawn from the reachable part of the belief simplex. One can derive a bound on the approximation error that is proportional to the sample spacing of $\mathcal{B}$ [Pineau et al., 2003]. The number of points required is driven by the curse of dimensionality: achieving a desired sample spacing requires a number of samples exponential in the dimensionality of the belief simplex.

However, in discounted problems, one can tolerate more approximation error at points that are only reachable after many time steps. This idea, which is not used in the sample spacing argument, is the basis of a second type of convergence result [Zhang and Zhang, 2001, Smith and Simmons, 2004], in which the error bound is derived from the fact that $\mathcal{B}$ samples enough of the search tree to some depth. The number of points required is driven by the curse of history: fully expanding the search tree to depth $t$ requires a number of points exponential in $t$.

This paper presents a new convergence argument that draws on both approaches. Our analysis applies to the case when the sample spacing varies according to what we call *discounted reachability*, which more accurately reflects the behavior of current algorithms (Fig. 1).

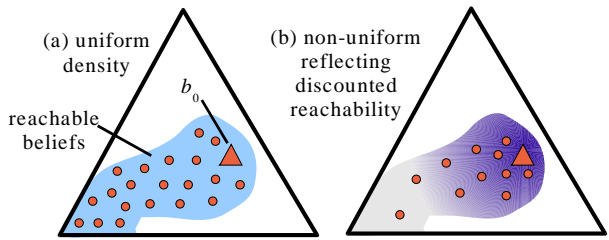

Figure 1: Sampling strategies for $\mathcal{B}$.

The remainder of the paper discusses recent improvements in our heuristic search value iteration algorithm (HSVI). HSVI is a point-based algorithm that maintains both upper and lower bounds on the optimal value function, allowing it to use effective heuristics for action and observation selection, and to provide provably small regret from the policy it generates [Smith and Simmons, 2004]. The new implementation of HSVI calculates tighter initial bounds, avoids solving linear programs, and makes more effective use of sparsity. Empirical results show speedups of more than two orders of magnitude.

## 2 POMDP INTRODUCTION

A POMDP models a planning problem that includes hidden state and uncertainty in action effects; the agent is assumed to know the transition model. Formally, a POMDP is described by a finite set of states $\mathcal{S} = \{s_1, \ldots, s_{|\mathcal{S}|}\}$, a finite set of actions $\mathcal{A} = \{a_1, \ldots, a_{|\mathcal{A}|}\}$, a finite set of observations $\mathcal{Z} = \{z_1, \ldots, z_{|\mathcal{Z}|}\}$, transition probabilities $T^{a,z}(s_i, s_j) = \Pr(s_j|s_i, a, z)$, a real-valued reward function $R(s, a)$, a discount factor $\gamma < 1$, and an initial belief $b_0$.

The Markov property of the model ensures that the agent can use a probability distribution $b$ over current states as a sufficient statistic for the history of actions and observations. Geometrically, the space of beliefs is a simplex, denoted $\Delta$. At each stage of forward simulation the current belief can be updated based on the latest action $a$ and observation $z$ using the formula $b' \leftarrow \tau(b, a, z)$, defined so that

$$b'(s') = \sum_s T^{a,z}(s, s')b(s) \quad (1)$$

Only a subset of $\Delta$ is reachable from $b_0$ through repeated applications of $\tau$; this subset is denoted $\bar{\Delta}$.

In general the object of the planning problem is to generate a policy $\pi$ that maximizes expected long-term reward:

$$J^\pi(b) = E\left[\sum_{t=0}^\infty \gamma^t R(s_t, a_t) \mid b, \pi\right] \quad (2)$$

A globally optimal policy $\pi^*$ is known to exist when $\gamma < 1$ [Howard, 1960]. We are particularly interested in the focused approximation setting, in which one attempts to generate a policy $\hat{\pi}$ that minimizes regret $J^{\pi^*}(b_0) - J^{\hat{\pi}}(b_0)$ when executed starting from $b_0$.

A POMDP is often solved by approximating its optimal value function $V^* = J^{\pi^*}$. Any value function $V$ induces a policy $\pi_V$ in which actions are selected via one-step lookahead. The regret of the policy induced by an approximate value function $\hat{V}$ can be forced arbitrarily small by reducing the max-norm error $\|V^* - \hat{V}\|_\infty$.

Value iteration starts with an initial guess $V_0$ and approximates $V^*$ through repeated application of the Bellman update $V_t \leftarrow HV_{t-1}$, where $H$ is defined as

$$HV(b) = \max_a \left[R(b, a) + \sum_{b'} \Pr(b'|b, a) V(b')\right] \quad (3)$$

$V^*$ satisfies Bellman's equation $V^* = HV^*$. When $\gamma < 1$, $H$ is a contraction and $V^*$ is the unique solution. During value iteration, each $V_t$ is piecewise linear and convex [Sondik, 1971], so it can be represented as a set of vectors $\Gamma_t = \{\alpha_1, \ldots, \alpha_{|\Gamma_t|}\}$, such that $V_t(b) = \max_i(\alpha_i \cdot b)$.

There are a number of value iteration algorithms that calculate $H$ exactly by projecting $\alpha$ vectors from $\Gamma_{t-1}$ to $\Gamma_t$ [Sondik, 1971, Cassandra et al., 1997]. Unfortunately, in the worst case the size of the representation grows as $|\Gamma_t| = |\mathcal{A}||\Gamma_{t-1}|^{|\mathcal{O}|}$, which rapidly becomes intractable even for modest problem sizes. Despite clever strategies for

---

**Algorithm 1.** $\beta = \text{backup}(\Gamma, b)$.

1. $\beta_{a,z} \leftarrow \text{argmax}_{\alpha \in \Gamma}(\alpha \cdot \tau(b, a, z))$
2. $\beta_a(s) \leftarrow R(s, a) + \gamma \sum_{s', z} \beta_{a,z}(s') T^{a,z}(s, s')$
3. $\beta \leftarrow \text{argmax}_{\beta_a}(\beta_a \cdot b)$

---

pruning dominated $\alpha$ vectors, these algorithms have been unable to scale to larger problems. The intractability of exact value iteration has led to the development of a wide variety of approximation techniques, too many to mention here [Aberdeen, 2002].

## 3 POINT-BASED ALGORITHMS

Point-based value iteration algorithms rely on the fact that performing many fast approximate updates often results in a more useful value function than performing a few exact updates. Their fundamental operation is the point-based update $\text{backup}(\Gamma, b)$, which generates a single $\alpha$ vector from $HV$ that is guaranteed to be maximal at $b$ (Alg. 1).

Our analysis focuses on a simple conceptual version of point-based value iteration. We assume there is a fixed finite set of beliefs $\mathcal{B}$. At each step the algorithm generates an $\alpha$ vector for every point in $\mathcal{B}$, and the set of vectors $\Gamma$ defines a value function through max-projection as described earlier. Denote the value function after $t$ updates as $V_t^\mathcal{B}$. The value function is initialized with $V_0^\mathcal{B} \leftarrow R_{\min}$, and the update rule is $V_t^\mathcal{B} \leftarrow H_\mathcal{B} V_{t-1}^\mathcal{B}$, where the update operator $H_\mathcal{B}$ applies the point-based update at every point of $\mathcal{B}$:

$$H_\mathcal{B} \Gamma = \{\text{backup}(\Gamma, b) \mid b \in \mathcal{B}\} \quad (4)$$

In this case, the approximation error relative to exact value iteration after $t$ updates, $\|V_t - V_t^\mathcal{B}\|_\infty$, is known to be bounded proportionally with the sample spacing $\delta(\mathcal{B})$, which is defined to be the maximum 1-norm distance from any point in $\bar{\Delta}$ to $\mathcal{B}$ [Pineau et al., 2003]. $\mathcal{B}$ thus needs to contain only enough points to cover $\bar{\Delta}$ with a uniform sample spacing.

However, current point-based algorithms do not sample $\bar{\Delta}$ uniformly (although the PBVI algorithm makes some attempt to do so). Instead, they collect points for $\mathcal{B}$ by a forward simulation process that biases $\mathcal{B}$ to contain beliefs that are only a few simulation steps away from $b_0$. However, this bias arguably helps rather than hurts. It underlies a second type of convergence argument based on the depth of the search tree. If $\mathcal{B}$ contains all the beliefs that result from expanding the search tree to depth $t$ and $t$ updates over $\mathcal{B}$ are performed, then the approximation error at $b_0$ is bounded proportionally with $\gamma^t$.

## 3.1 NEW THEORETICAL RESULTS

This section presents a new convergence argument that draws on the two earlier approaches. Its use of weighted max-norm machinery in value iteration is closely related to [Munos, 2004]. Our argument reflects current point-based algorithms in that it allows $\mathcal{B}$ to be a non-uniform sampling of $\bar{\Delta}$ whose spacing varies according to *discounted reachability*. The discounted reachability $\rho : \bar{\Delta} \to \mathbb{R}$ is defined to be $\rho(b) = \gamma^L$, where $L$ is the length of the shortest sequence of belief-state transitions from $b_0$ to $b$. $\rho$ satisfies the property that $\rho(b') \geq \gamma \rho(b)$ whenever there is a single-step transition from $b$ to $b'$. Based on $\rho$, we define a generalized sample spacing measure $\delta_p$ (with $0 \leq p < 1$):

$$\delta_p(\mathcal{B}) = \max_{b \in \bar{\Delta}} \min_{b' \in \mathcal{B}} \frac{\|b - b'\|_1}{\rho^p(b)} \quad (5)$$

In order to achieve a small $\delta_p$ value, $\mathcal{B}$ must have small 1-norm distance from all points in $\bar{\Delta}$, but its distance from $b$ can be proportionally larger if $\rho^p(b)$ is small.

When sample spacing is bounded in terms of $\delta_p$, $H_\mathcal{B}$ does not have the error properties we want under the usual max-norm $\|\cdot\|_\infty$. We must define a new norm to reflect the fact that $H_\mathcal{B}$ induces larger errors where $\rho$ is small. A weighted max-norm is a function $\|\cdot\|_\xi$ such that

$$\|V - \bar{V}\|_\xi = \max_b \frac{|V(b) - \bar{V}(b)|}{\xi(b)}, \quad (6)$$

where $\xi > 0$. Not surprisingly, $\|\cdot\|_{\rho^p}$ is the norm we need.

Note that when $p = 0$, $\delta_p$ reduces to the uniform spacing measure $\delta$ and $\|\cdot\|_{\rho^p}$ reduces to the max-norm. We begin by generalizing some well-known results about standard value iteration to the $\rho^p$-norm with $0 \leq p < 1$.

**Theorem 1.** *The exact Bellman update $H$ is a contraction under the $\rho^p$-norm with contraction factor $\gamma^{1-p}$.*

*Proof.* Define

$$Q_a^V(b) = R(b,a) + \gamma \sum_{b'} \Pr(b'|b,a) V(b') \quad (7)$$

so that $HV = \max_a Q_a^V$. For any $a$, the mapping $V \mapsto Q_a^V$ has contraction factor $\gamma^{1-p}$:

$$\|Q_a^V - Q_a^{\bar{V}}\|_{\rho^p} = \max_b \frac{|Q_a^V(b) - Q_a^{\bar{V}}(b)|}{[\rho(b)]^p} \quad (8)$$

$$= \max_b \gamma \sum_{b'} \Pr(b' \mid b, a) \frac{|V(b') - \bar{V}(b')|}{[\rho(b)]^p} \quad (9)$$

$$\leq \max_b \gamma \sum_{b'} \Pr(b'|b,a) \frac{|V(b') - \bar{V}(b')|}{[\gamma \rho(b')]^p} \quad (10)$$

$$\leq \max_b \gamma^{1-p} \sum_{b'} \Pr(b'|b,a) \|V - \bar{V}\|_{\rho^p} \quad (11)$$

$$= \gamma^{1-p} \|V - \bar{V}\|_{\rho^p} \quad (12)$$

Now choose an arbitrary $b \in \bar{\Delta}$. Assume without loss of generality that $HV(b) \geq H\bar{V}(b)$. Choose $a^*$ to maximize $Q_{a^*}^V(b)$, and $\bar{a}$ to maximize $Q_{\bar{a}}^{\bar{V}}(b)$. It follows that $Q_{a^*}^{\bar{V}}(b) \leq Q_{\bar{a}}^{\bar{V}}(b) \leq Q_{a^*}^V(b)$, and

$$|HV(b) - H\bar{V}(b)| = |Q_{a^*}^V(b) - Q_{\bar{a}}^{\bar{V}}(b)| \quad (13)$$

$$\leq |Q_{a^*}^V(b) - Q_{a^*}^{\bar{V}}(b)| \quad (14)$$

$$\leq \max_a |Q_a^V(b) - Q_a^{\bar{V}}(b)| \quad (15)$$

Dividing through by $\rho^p(b)$ and maximizing over $b$ yields

$$\|HV - H\bar{V}\|_{\rho^p} \leq \max_a \|Q_a^V(b) - Q_a^{\bar{V}}(b)\|_{\rho^p} \quad (16)$$

$$\leq \gamma^{1-p} \|V - \bar{V}\|_{\rho^p} \quad \square$$

**Theorem 2.** *Let $\hat{\pi}$ be the one-step lookahead policy induced by an approximate value function $\hat{V}$. The regret from executing $\hat{\pi}$ rather than $\pi^*$, starting from $b_0$, is at most*

$$\frac{2\gamma^{1-p}}{1 - \gamma^{1-p}} \|V^* - \hat{V}\|_{\rho^p} \quad (17)$$

*Proof.* Choose an arbitrary $b \in \bar{\Delta}$. It is easy to check that for any policy $\pi$, $J^\pi(b) = Q_{\pi(b)}^{J^\pi}(b)$. Also, because $\hat{\pi}$ is the one-step lookahead policy induced by $\hat{V}$, $Q_{\hat{\pi}(b)}^{\hat{V}}(b) = H\hat{V}(b)$. The Bellman equation states that $V^* = HV^*$. Then:

$$|J^{\pi^*}(b) - J^{\hat{\pi}}(b)| = |V^*(b) - Q_{\hat{\pi}(b)}^{J^{\hat{\pi}}}(b)| \quad (18)$$

$$= |V^*(b) - Q_{\hat{\pi}(b)}^{\hat{V}}(b) + Q_{\hat{\pi}(b)}^{\hat{V}}(b) - Q_{\hat{\pi}(b)}^{J^{\hat{\pi}}}(b)| \quad (19)$$

$$\leq |V^*(b) - Q_{\hat{\pi}(b)}^{\hat{V}}(b)| + |Q_{\hat{\pi}(b)}^{\hat{V}}(b) - Q_{\hat{\pi}(b)}^{J^{\hat{\pi}}}(b)| \quad (20)$$

$$\leq |HV^*(b) - H\hat{V}(b)| +$$
$$\gamma \sum_{b'} \Pr(b'|b, \hat{\pi}(b)) \, |\hat{V}(b') - J^{\hat{\pi}}(b')| \quad (21)$$

$$\leq |HV^*(b) - H\hat{V}(b)| +$$
$$\gamma \sum_{b'} \Pr(b'|b, \hat{\pi}(b)) \, \gamma^{-p} \rho^p(b) \|\hat{V} - J^{\hat{\pi}}\|_{\rho^p} \quad (22)$$

$$\leq |HV^*(b) - H\hat{V}(b)| + \gamma^{1-p} \rho^p(b) \|\hat{V} - J^{\hat{\pi}}\|_{\rho^p} \quad (23)$$

Dividing through by $\rho^p(b)$ and maximizing over $b$ gives

$$\|J^{\pi^*} - J^{\hat{\pi}}\|_{\rho^p} \quad (24)$$

$$\leq \|HV^* - H\hat{V}\|_{\rho^p} + \gamma^{1-p} \|\hat{V} - J^{\hat{\pi}}\|_{\rho^p} \quad (25)$$

$$\leq \gamma^{1-p} \big( \|V^* - \hat{V}\|_{\rho^p} + \|\hat{V} - J^{\hat{\pi}}\|_{\rho^p} \big) \quad (26)$$

$$\leq \gamma^{1-p} \big( \|V^* - \hat{V}\|_{\rho^p} + \quad (27)$$
$$\|\hat{V} - V^*\|_{\rho^p} + \|V^* - J^{\hat{\pi}}\|_{\rho^p} \big) \quad (28)$$

$$\leq \gamma^{1-p} \big( 2\|V^* - \hat{V}\|_{\rho^p} + \|V^* - J^{\hat{\pi}}\|_{\rho^p} \big) \quad (29)$$

$$= \gamma^{1-p} \big( 2\|V^* - \hat{V}\|_{\rho^p} + \|J^{\pi^*} - J^{\hat{\pi}}\|_{\rho^p} \big) \quad (30)$$

$$= 2\gamma^{1-p} \|V^* - \hat{V}\|_{\rho^p} + \gamma^{1-p} \|J^{\pi^*} - J^{\hat{\pi}}\|_{\rho^p} \quad (31)$$

Solving the recursion,

$$\|J^{\pi^*} - J^{\hat{\pi}}\|_{\rho^p} \leq \frac{2\gamma^{1-p}}{1-\gamma^{1-p}}\|V^* - \hat{V}\|_{\rho^p} \quad (32)$$

And since $\rho(b_0) = 1$, we have the desired regret bound:

$$J^{\pi^*}(b_0) - J^{\hat{\pi}}(b_0) \leq \frac{2\gamma^{1-p}}{1-\gamma^{1-p}}\|V^* - \hat{V}\|_{\rho^p} \quad \square$$

It is worth noting (although we lack space to prove it here) that a tighter bound applies when $\hat{V}$ is uniformly improvable [Zhang and Zhang, 2001]. A small modification to $H_\mathcal{B}$ would make $V_t^\mathcal{B}$ uniformly improvable at the cost of increasing $|\Gamma|$. In that case the regret would be at most $\gamma^{1-p}\|V^* - \hat{V}\|_{\rho^p}$.

Having discussed the $\rho^p$-norm behavior of $H$, now we move on to the $\rho^p$-norm behavior of $H_\mathcal{B}$ with non-uniform sample spacing $\delta_p$.

**Lemma 1.** *At any update step $t$, the error $\|HV_t^\mathcal{B} - H_\mathcal{B}V_t^\mathcal{B}\|_{\rho^p}$ introduced by a single application of $H_\mathcal{B}$ rather than $H$ is at most*

$$\frac{(R_{\max} - R_{\min})\delta_p(\mathcal{B})}{1-\gamma^{1-p}} \quad (33)$$

*Proof.* The argument is analogous to Lemma 1 of [Pineau et al., 2003]. Necessary changes: (1) restrict $b'$ to be drawn from $\bar{\Delta}$, (2) divide throughout by $\rho^p(b')$, and (3) substitute $\gamma^{1-p}$ for $\gamma$ in the denominator to reflect the changed contraction properties of $H$ under the new norm. $\square$

**Theorem 3.** *At any update step $t$, the accumulated error $\|V_t - V_t^\mathcal{B}\|_{\rho^p}$ is at most*

$$\frac{(R_{\max} - R_{\min})\delta_p(\mathcal{B})}{(1-\gamma^{1-p})^2} \quad (34)$$

*Proof.* The argument is analogous to Theorem 1 of [Pineau et al., 2003]. Necessary changes: (1) replace the max-norm with the $\rho^p$-norm, and (2) replace $\gamma$ with $\gamma^{1-p}$. $\square$

Taken together, these results show that the conceptual algorithm can be used to generate a policy with arbitrarily small regret related to $\delta_p(\mathcal{B})$, and they provide a finite bound on the number of updates required to achieve a given regret.

### 3.2 IMPLICATIONS FOR ALGORITHM DESIGN

The bias of our model toward beliefs with high discounted reachability describes current algorithms more accurately than uniform sampling, at least to the extent that the algorithms perform a (typically shallow) forward exploration from the initial belief to generate $\mathcal{B}$.

The parameter $p$ arose naturally during our analysis. $p = 0$ corresponds to uniform sampling and the usual max-norm. As $p$ increases, samples grow less dense in areas with low reachability and the norm becomes correspondingly more tolerant. But the results show that there's no free lunch: the higher effective discount factor $\gamma^{1-p}$ under the new norm means that more updates are required and the final error bounds are looser. The new theoretical framework provides a way to analyze this trade-off.

We initially found the concept of discounted reachability surprising. The intuition is that (1) beliefs that are deeper in the search tree are less relevant, and (2) beliefs that can only be reached by low-probability belief transitions are less relevant. But discounted reachability ignores (2) entirely, in that all transitions with non-zero probability are treated equally.

Actually, we started with a different concept of *discounted occupancy*, in which beliefs are tagged as proportionally less relevant if they can only be reached by low-probability belief transitions. The bias of current algorithms seems to be better described by discounted occupancy, and empirically, treating all transitions with non-zero probability equally hurts performance. But the convergence results we found do *not* go through when discounted occupancy is used instead of discounted reachability. We hope that a more sophisticated future analysis will shed light on this issue.

In summary, these new results take us closer to understanding point-based algorithms. The analysis helps explain important trade-offs in algorithm design, although we have not yet had time to apply it to a working algorithm. The next section changes the topic to recent improvements in our (point-based) HSVI algorithm. Note that those improvements are not based on the theoretical results just presented.

## 4 IMPROVEMENTS IN HEURISTIC SEARCH VALUE ITERATION

This section discusses recent improvements in our heuristic search value iteration algorithm (HSVI). Relative to our original presentation of HSVI, the new implementation calculates tighter initial bounds, avoids solving linear programs, and makes more effective use of sparsity. Empirical results show speedups of up to three orders of magnitude.

### 4.1 HSVI OVERVIEW

HSVI is a point-based algorithm that maintains both upper and lower bounds on the optimal value function, allowing it to use effective heuristics for action and observation selection, and to provide provably small regret from the policy it generates. We provide a brief overview here;

**Algorithm 2.** $\pi = \text{HSVI}(\epsilon)$.

HSVI($\epsilon$) returns a policy $\pi$ whose regret relative to $\pi^*$, starting from $b_0$, is at most $\epsilon$.

1. Initialize the bounds $\hat{V}$.
2. While $\text{width}(\hat{V}(b_0)) > \epsilon$, repeatedly invoke $\text{explore}(b_0, \epsilon, 0)$.
3. Having achieved the desired precision, return the direct-control policy $\pi$ corresponding to the lower bound.

**Algorithm 3.** $\text{explore}(b, \epsilon, t)$.

explore recursively follows a single path down the search tree until satisfying a termination condition based on the width of the bounds interval. It then performs a series of updates on its way back up to the initial belief.

1. If $\text{width}(\hat{V}(b)) \leq \epsilon \gamma^{-t}$, return.
2. Select an action $a^*$ and observation $z^*$ according to the forward exploration heuristics.
3. Call $\text{explore}(\tau(b, a^*, z^*), \epsilon, t+1)$.
4. Perform a point-based update of $\hat{V}$ at belief $b$.

for more detail refer to [Smith and Simmons, 2004]. We refer to the original version and our current version as HSVI1 and HSVI2, respectively. The differences are covered comprehensively in §4.2.

HSVI is outlined in Algs. 2 and 3. We denote the lower and upper bound functions as $\underline{V}$ and $\bar{V}$, respectively. The interval function $\hat{V}$ refers to them collectively, such that $\hat{V}(b) = [\underline{V}(b), \bar{V}(b)]$ and $\text{width}(\hat{V}(b)) = \bar{V}(b) - \underline{V}(b)$.

### 4.1.1 Value Function Representation

HSVI uses the usual $\Gamma$ vector set representation for its lower bound (see §2). Unfortunately, if the upper bound is represented with a vector set, updating by adding a vector does not have the desired effect of improving the bound in the neighborhood of the local update. To accommodate the need for updates, HSVI uses a *point set* representation for the upper bound. The value at a point $b$ is the projection of $b$ onto the convex hull formed by a finite set $\Upsilon$ of belief/value points $(b_i, \bar{v}_i)$. Updates are performed by adding a new point to the set. In HSVI1, the projection onto the convex hull is calculated by solving a linear program using the commercial CPLEX software package.

### 4.1.2 Initialization

HSVI1 initializes the lower bound using a conservative estimate of the values of blind policies of the form "always execute action $a$". The smallest possible reward from executing action $a$ is $\min_s R(s, a)$, so a bound on the long-term reward for that policy can be found by evaluating the relevant summation. HSVI1 then maximizes over $a$:

$$\underline{R} \leftarrow \max_a \sum_{t=0}^{\infty} \gamma^t \min_s R(s, a) = \frac{\max_a \min_s R(s, a)}{1 - \gamma} \quad (35)$$

The vector set for the initial lower bound $\underline{V}_0$ contains a single vector $\alpha$ such that every $\alpha(s) = \underline{R}$.

HSVI1 initializes the upper bound by assuming full observability and solving the MDP version of the problem. This provides upper bound values at the corners of the belief simplex, which form the initial point set.

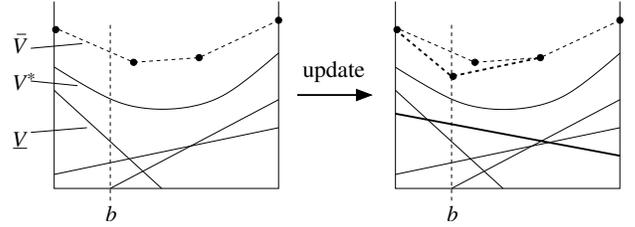

Figure 2: Local update at $b$.

### 4.1.3 Local Updates

HSVI performs a local update $L_b$ of the lower bound by adding the result of a point-based update at $b$ to the vector set:

$$L_b \Gamma = \Gamma \cup \{\text{backup}(\Gamma, b)\} \quad (36)$$

It performs a local update $U_b$ of the upper bound by adding the result of a Bellman update at $b$ to the point set:

$$U_b \Upsilon = \Upsilon \cup \{(b, H\bar{V}(b))\} \quad (37)$$

Fig. 2 represents the structure of the bounds representations and the process of locally updating at $b$. In the left side of the figure, the points and dotted lines represent $\bar{V}$ (upper bound points and convex hull). Several solid lines represent the vectors of $\Gamma$. In the right side of the figure, we see the result of updating both bounds at $b$, which involves adding a new point to $\Upsilon$ and a new vector to $\Gamma$, bringing both bounds closer to $V^*$.

HSVI periodically prunes dominated elements in both the lower bound vector set and the upper bound point set; we do not discuss the pruning here because it is unaffected by our recent changes.

### 4.1.4 Forward Exploration Heuristics

This section discusses the heuristics that are used to decide which child of the current node to visit as HSVI works its way forward from the initial belief. Starting from parent node $b$, HSVI must choose an action $a^*$ and an observation $z^*$: the child node to visit is $\tau(b, a^*, z^*)$.

HSVI selects actions greedily based on the upper bound (the IE-MAX heuristic). At a belief $b$, for every action, it can compute an upper bound on the long-term reward from taking that action. It chooses the action with the highest upper bound:

$$a^* = \operatorname*{argmax}_{a} Q_a^{\bar{V}}(b) \qquad (38)$$

Because the bounds at a parent node are always wider than the bounds at the child with the highest upper bound value, choosing $a^*$ according to IE-MAX is a good way to ensure convergence. In the simpler context where updates do not affect neighboring nodes, it is provably optimal [Kaelbling, 1993].

HSVI uses the *weighted excess uncertainty* heuristic for observation selection. Excess uncertainty at a belief $b$ with depth $t$ in the search tree is defined to be

$$\operatorname{excess}(b, t) = \operatorname{width}(\hat{V}(b)) - \epsilon \gamma^{-t} \qquad (39)$$

Excess uncertainty has the property that if all the children of a node $b$ have negative excess uncertainty, then after an update $b$ will also have negative excess uncertainty. Negative excess uncertainty at the root implies the desired convergence to within $\epsilon$.

The weighted excess uncertainty heuristic is designed to focus attention on the child node with the greatest contribution to the excess uncertainty at the parent:

$$z^* = \operatorname*{argmax}_{z} \left[ \Pr(z|b, a^*) \operatorname{excess}(\tau(b, a^*, z), t+1) \right] \qquad (40)$$

Both the action and observation selection heuristics are designed so that applying them systematically guarantees HSVI convergence in finite time [Smith and Simmons, 2004].

## 4.2 CHANGES BETWEEN HSVI1 AND HSVI2

We report a series of changes made since our initial presentation of HSVI1. The changes are roughly ordered in terms of their impact on the overall performance. The relative speedup for individual changes is problem-dependent; the reported values were measured informally on the *Tag* problem. HSVI2 performance is presented in §4.3.

### 4.2.1 More Effective Use of Sparsity

HSVI1 represents beliefs and transition functions as vectors and matrices in BLAS compressed storage mode [Dongarra et al., 1988]. It uses an off-the-shelf sparse linear algebra package to compute belief transitions and take dot products. That package turned out to be using inappropriate algorithms, slowing down individual operations by as much as 100x. We addressed the problem in HSVI2 by writing our own simple compressed storage operations, which speed up lower bound updates by about 50x.

HSVI1 represents $\alpha$ vectors in dense storage mode because they tend to have a large number of non-zeros, even when beliefs are sparse. Typically, when $\alpha \leftarrow \operatorname{backup}(\Gamma, b)$ is applied, all of the entries of $\alpha$ must be computed, even if $b$ is sparse and most of the entries have no effect on the value $\alpha \cdot b$. They are required because HSVI may later need to evaluate $\alpha \cdot b'$ where $b'$ has different non-zeros.

But if $\alpha$ is optimized for $b$, why should we expect it to be relevant to $b'$, which has different non-zeros and perhaps no overlap with $b$ at all? This leads to the idea of *masked* $\alpha$ vectors. In HSVI2, $\alpha \leftarrow \operatorname{backup}(\Gamma, b)$ computes only the entries of $\alpha$ that correspond to non-zeros of $b$. A mask records which entries were computed. If HSVI2 later evaluates $\max_i(\alpha_i \cdot b')$ and $b'$ has a non-zero in a position that was not computed in $\alpha_i$, the dot product $\alpha_i \cdot b'$ is rejected from consideration.

This change can be interpreted geometrically. Sparse beliefs lie in hyperplanes on the boundary of the belief simplex. When a masked $\alpha$ vector is computed using the new $\operatorname{backup}(\Gamma, b)$, it applies only to the lowest-dimensional boundary hyperplane containing $b$. Empirically, masked $\alpha$ vectors speed up lower bound updates by about 5x. Note that almost any POMDP value iteration algorithm could make use of this concept.

### 4.2.2 Avoid Solving Linear Programs

HSVI1 evaluates $\bar{V}(b)$ by computing the exact projection of $b$ onto the convex hull of the points in $\Upsilon$, which involves solving a linear program with the commercial CPLEX software package. Each upper bound update requires several such projections, and the time spent solving linear programs dominates the upper bound update time.

HSVI2 uses an approximate projection onto the convex hull suggested by [Hauskrecht, 2000]. Projection onto the convex hull of a set of points is particularly simple when the set contains only the corners of the belief simplex and one interior point: it can be computed in $O(|\mathcal{S}|)$ time. To approximately project onto the overall convex hull, HSVI2 runs this operation for each interior point of $\Upsilon$ and takes the minimum value, requiring $O(|\Upsilon||\mathcal{S}|)$ time overall (or less with sparsity). This approximate convex hull has the key properties that (1) it is everywhere greater than the exact convex hull, and (2) the approximation at $b$ is exact if there is an undominated pair $(b, \bar{v}) \in \Upsilon$. Empirically, the approximate projection speeds up upper bound updates by about 100x.

### 4.2.3 Tighter Initial Bounds

HSVI1 generates an initial lower bound based on a conservative estimate of the values of blind policies. HSVI2 uses a better blind policy value estimate suggested in [Hauskrecht, 1997]. The value $\alpha^a$ of each policy "always

take action $a$" is updated in MDP fashion:

$$\alpha_{t+1}^a(s) = R(s,a) + \gamma \sum_{s'} \Pr(s'|s,a)\alpha_t^a(s') \qquad (41)$$

Each update of $|\mathcal{A}|$ vectors can be evaluated in $O(|\mathcal{S}|^2|\mathcal{A}|)$ time. HSVI2 initializes the vectors $\alpha_0^a$ using the HSVI1 lower bound, which guarantees that the bound is valid even if the iteration is not run to completion. When the iteration is stopped, the $\alpha_t^a$ vectors form the initial lower bound $\Gamma$.

HSVI1 generates an initial upper bound based on the value function of the fully observable MDP. HSVI2 uses the fast informed bound (FIB) approximation, which is guaranteed to give a tighter upper bound than the MDP approximation [Hauskrecht, 2000]. FIB iteration keeps one vector $\alpha^a$ for each action and uses the following update rule:

$$\alpha_{t+1}^a(s) = R(s,a) + \gamma \sum_z \max_{a'} \sum_{s'} \Pr(s',z|s,a)\alpha_t^{a'}(s')$$

Each FIB update can be evaluated in $O(|\mathcal{A}|^2|\mathcal{S}|^2|\mathcal{Z}|)$ time. As with the lower bound, HSVI2 initializes the upper bound vectors $\alpha_0^a$ using the HSVI1 upper bound. When FIB iteration is stopped, each corner point corresponding to a state $s$ is initialized to $\max_a \alpha_t^a(s)$.

Empirically, HSVI2 can run both bound initialization routines to approximate convergence (residual $< 10^{-3}$) in at most a few seconds for all of the problems in our benchmark set. This results in better performance near the beginning of HSVI2 execution, although later in the run the effect is less significant. The change in the lower bound initialization is the more important of the two; the MDP upper bound was already fairly good for most problems.

### 4.3 HSVI2 PERFORMANCE

Fig. 3 shows HSVI2 reward vs. time for four problems from the scalable POMDP literature. The plotted reward is the average received over 100 or more simulations. We also plot HSVI2's bounds $\underline{V}(b_0)$ and $\bar{V}(b_0)$. HSVI2 was run only once on each problem since it is not stochastic. The platform used was a Pentium-4 running at 3.4 GHz, with 2 GB of RAM (HSVI2 used at most 250 MB of RAM).

The plots show a range of behaviors. *RockSample*[4,4] is especially easy; the HSVI2 bounds converge after 13 seconds, showing that the solution is optimal. *Hallway2* shows HSVI2 quickly arriving at an apparently near-optimal solution, but its bounds remain loose. *Tag* and *RockSample*[10,10] show typical behavior for large problems: the upper bound decrease is slow and steady while the lower bound (and the reward) improve in jumps, plateauing for long periods. *RockSample*[10,10], with $> 10^5$ states, would be too large for most POMDP algorithms to handle; HSVI2 gains by use of sparsity. It would run out of memory with a problem 5-10 times larger.

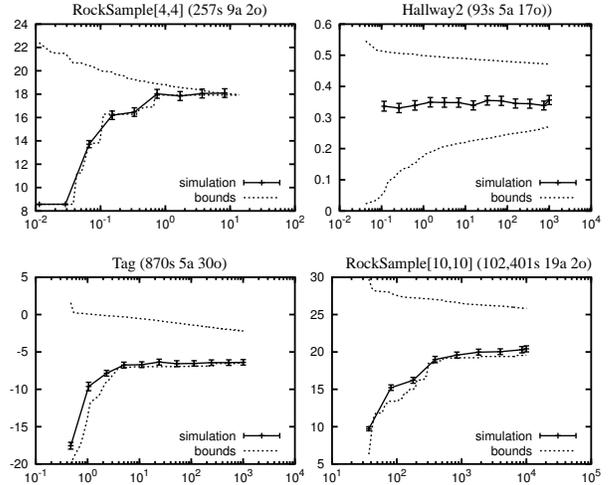

Figure 3: HSVI2 reward vs. wallclock time.

Fig. 4 shows running times and solution quality for HSVI and several other algorithms. Note that different algorithms were run on different platforms, so running times are only roughly comparable. The table also shows, for each problem, the 95% confidence interval for reward measurements assuming the variance of HSVI2's best policy and averaging 100 rewards. An algorithm's reward is starred if it is within the confidence interval relative to the best reported value.

HSVI2 is within measurement error of the best reported reward for all problems, and its running time is considerably shorter than other algorithms in most cases. The greatest speedup from HSVI1 to HSVI2 was observed on the *Rock-Sample*[7,8] problem. HSVI2 takes about 6 seconds to surpass the reward reached by HSVI1 after $> 10^4$ seconds. After correcting for running on a processor about 5x faster, this is $> 300$x speedup.

Other state-of-the-art scalable POMDP algorithms could not be compared to HSVI2 because they were tested on different problems. Among these, two techniques appear especially promising. Exponential-family PCA transforms the POMDP, compresses to a low-dimensional representation in the transformed space, then solves it with a grid-based algorithm. It has demonstrated good results on large-scale robot navigation problems [Roy and Gordon, 2003].

Value-directed compression (VDC) is another compression technique. It typically produces a less compact representation than E-PCA, but the compressed POMDP retains linear structure and value function convexity, so that it can be solved using almost any POMDP algorithm. The combinations VDC+BPI and VDC+PBVI have demonstrated scalability to huge problem sizes, up to 33 million states [Poupart and Boutilier, 2004].[1] VDC would likely boost

---
[1]VDC+PBVI results courtesy of Poupart, personal communi-

| Problem (states/actions/observations) | Reward | Time (s) | $|\Gamma|$ |
|---|---|---|---|
| **Tiger-Grid** (36s 5a 17o) | ($\pm 0.14$) | | |
| HSVI1 [Smith et al., 2004] | 2.35* | 10341 | 4860 |
| Perseus [Spaan et al., 2004] | 2.34* | 104 | 134 |
| HSVI2 | 2.30* | 52 | 1003 |
| PBUA [Poon, 2001] | 2.30* | 12116 | 660 |
| PBVI [Pineau et al., 2003] | 2.25* | 3448 | 470 |
| BPI [Poupart et al., 2003] | 2.22* | 1000 | 120 |
| QMDP | 0.26 | 0.026 | N/A |
| **Hallway** (61s 5a 21o) | ($\pm 0.038$) | | |
| PBVI [Pineau et al., 2003] | 0.53* | 288 | 86 |
| PBUA [Poon, 2001] | 0.53* | 450 | 300 |
| HSVI2 | 0.52* | 2.4 | 147 |
| HSVI1 [Smith et al., 2004] | 0.52* | 10836 | 1341 |
| Perseus [Spaan et al., 2004] | 0.51* | 35 | 55 |
| BPI [Poupart et al., 2003] | 0.51* | 185 | 43 |
| QMDP | 0.14 | 0.012 | N/A |
| **Hallway2** (93s 5a 17o) | ($\pm 0.048$) | | |
| HSVI2 | 0.35* | 1.5 | 114 |
| Perseus [Spaan et al., 2004] | 0.35* | 56 | 10 |
| HSVI1 [Smith et al., 2004] | 0.35* | 10010 | 1571 |
| PBUA [Poon, 2001] | 0.35* | 27898 | 1840 |
| PBVI [Pineau et al., 2003] | 0.34* | 360 | 95 |
| BPI [Poupart et al., 2004] | 0.32* | 790 | 60 |
| QMDP | 0.052 | 0.02 | N/A |
| **Tag** (870s 5a 30o) | ($\pm 1.2$) | | |
| Perseus [Spaan et al., 2004] | -6.17* | 1670 | 280 |
| HSVI2 | -6.36* | 24 | 415 |
| HSVI1 [Smith et al., 2004] | -6.37* | 10113 | 1657 |
| BPI [Poupart et al., 2004] | -6.65* | 250 | 17 |
| PBVI [Pineau et al., 2003] | -9.18 | 180880 | 1334 |
| QMDP | -16.48 | 0.07 | N/A |
| **RockSample[4,4]** (257s 9a 2o) | ($\pm 1.2$) | | |
| HSVI2 | 18.0* | 0.75 | 177 |
| HSVI1 [Smith et al., 2004] | 18.0* | 577 | 458 |
| PBVI [Pineau, pers. communication] | 17.1* | ~2000 | ? |
| QMDP | 3.5 | 0.008 | N/A |
| **RockSample[7,8]** (12,545s 13a 2o) | ($\pm 1.2$) | | |
| HSVI2 | 20.6* | 1003 | 2491 |
| HSVI1 [Smith et al., 2004] | 15.1 | 10266 | 94 |
| QMDP | 0 | 2.1 | N/A |
| **RockSample[10,10]** (102,401s 19a 2o) | ($\pm 1.3$) | | |
| HSVI2 | 20.4* | 10014 | 3199 |
| QMDP | 0 | 57 | N/A |

Figure 4: Multi-algorithm performance comparison.

HSVI scalability in a similar way.

## 5 CONCLUSION

We presented new theoretical results for point-based algorithms, which combine curse of dimensionality and curse of history arguments into an overall bound on the convergence of point-based value iteration with non-uniform sample spacing. In the future we will apply these results to point-based algorithm design.

We also demonstrated improved performance for our HSVI algorithm, with speedups of more than two orders of magnitude and successful scaling to a POMDP with $> 10^5$ states. In the future we would like to combine HSVI with a compact representation technique such as VDC to deal with still larger problems.

**Acknowledgments**

Thanks to Geoff Gordon and Pascal Poupart for helpful discussions. This work was funded in part by a NASA GSRP Fellowship with Ames Research Center.

cation.